\documentclass{article}

\usepackage{arxiv}

\usepackage[utf8]{inputenc} 
\usepackage[T1]{fontenc}    
\usepackage{hyperref}       
\usepackage{url}            
\usepackage{booktabs}       
\usepackage{amsfonts}       
\usepackage{nicefrac}       
\usepackage{microtype}      
\usepackage{lipsum}
\usepackage{amsmath}
\usepackage{graphicx}
\usepackage{amssymb}
\usepackage{subfig}

\title{PaddleSeg: A High-Efficient Development Toolkit for Image Segmentation}

\author{
  Yi Liu\\
  Baidu Inc.\\
  \texttt{liuyi22@baidu.com} \\
    \And
 Lutao Chu \\
  Baidu Inc.\\
  \texttt{chulutao@baidu.com} \\
  \And
  Guowei Chen \\
  Baidu Inc. \\
  \texttt{chenguowei01@baidu.com} \\
    \And
  Zewu Wu \\
  Baidu Inc. \\
  \texttt{wuzewu@baidu.com} \\
  
  \And
  Zeyu Chen \\
  Baidu Inc. \\
  \texttt{chenzeyu01@baidu.com} \\
      \And
  Baohua Lai \\
  Baidu Inc. \\
  \texttt{laibaohua@baidu.com} \\
     \And
  Yuying Hao \\
  Baidu Inc. \\
  \texttt{haoyuying@baidu.com} \\

}

\begin{document}
\maketitle

\begin{abstract}
Image Segmentation plays an essential role in computer vision and image processing with various applications from medical diagnosis to autonomous car driving. A lot of segmentation algorithms have been proposed for addressing specific problems. In recent years, the success of deep learning techniques has tremendously influenced a wide range of computer vision areas, and the modern approaches of image segmentation based on deep learning are becoming prevalent. In this article, we introduce a high-efficient development toolkit for image segmentation, named PaddleSeg. The toolkit aims to help both developers and researchers in the whole process of designing segmentation models, training models, optimizing performance and inference speed, and deploying models. Currently, PaddleSeg supports around 20 popular segmentation models and more than 50 pre-trained models from real-time and high-accuracy 
levels. With modular components and backbone networks, users can easily build over one hundred models for different requirements. Furthermore, we provide comprehensive benchmarks and evaluations to show that these segmentation algorithms trained on our toolkit have more competitive accuracy. Also, we provide various real industrial applications and practical cases based on PaddleSeg.  All codes and examples of PaddleSeg are available at \url{https://github.com/PaddlePaddle/PaddleSeg}.
\end{abstract}


\section{Introduction}
Image Segmentation is one of the fast-growing areas in computer vision and image understanding. It plays a crucial role in diverse applications such as industrial inspection, medical image diagnosis, autonomous driving car, satellite image processing, and human body parsing~\cite{takos2020survey,minaee2020image,lateef2019survey}, etc. Image Segmentation involves partitioning an image into multiple segments through labelling every pixel in the image. In general, it is also regarded as a pixel-level classification problem, which requires much higher accuracy than image-level classification~\cite{krizhevsky2017imagenet} or object-level detection~\cite{zhao2019object}. However, there are critical issues in image segmentation. The number of segmentation labels is very limited, and the annotation task is quite time-consuming, e.g. annotation and quality control required more than 1.5 h for a single image in the Cityscapes dataset~\cite{cordts2016cityscapes}. The complex and diverse scenes in nature also increase the level of difficulty on image segmentation. Therefore, image segmentation is a long-standing challenge in computer vision, and numerous great algorithms have been proposed to solve the different segmentation problems.


In an early attempt at image segmentation, thresholding is widely used to divide an image into the object region and the background region~\cite{al2010image}. Due to its simplicity, it works well even on a single gray-scale image. Then clustering methods are also popular in the early stage~\cite{ozden2005image}, which involves partitioning the image into K groups according to their similarity on color, gradient or relative distance, etc. Edge detection is another type of segmentation, which describes the boundaries of different regions~\cite{muthukrishnan2011edge}. Later, graph-based and conditional random fields (CRFs) approaches process the image as a graph, which minimize the cost of graph cut~\cite{peng2009iterated,lafferty2001conditional}, and until now these efficient methods are still usually taken as a post-processing method in some more modern algorithms. Despite the popularity of those kinds of methods, the challenging segmentation problems have yet to be resolved properly and efficiently.

Over the past few years, with the significant success of deep learning in visual recognition tasks, image segmentation has entered a new era. Deep neural networks have remarkable performance improvements and often achieve the highest accuracy rates on popular benchmarks~\cite{minaee2020image,deng2009imagenet, everingham2010pascal, lin2014microsoft, cordts2016cityscapes}. In the recent literature, a lot of segmentation algorithms based on deep neural networks have sprung up. Convolution neural networks (CNNs) were initially used for classification tasks~\cite{krizhevsky2017imagenet, he2016deep}. Nowadays, fully convolutional networks (FCNs) have been becoming the most extensive architectures in the image segmentation filed~\cite{long2015fully}. 

In this article, we proposed PaddleSeg, which is a high-efficient development toolkit for image segmentation. The toolkit aims to accelerate the image segmentation progress in both industry and academic research. It can help developers and researchers in the whole development process, such as designing segmentation models, training models, optimizing model performance, improving inference speed, and deploying models. To the best of our knowledge, PaddleSeg is the only one toolkit that supports such an end-to-end development process in image segmentation.

PaddleSeg has implemented around twenty high-quality models from real-time to high-accuracy levels, such as Fast-SCNN~\cite{poudel2019fast}, BiSeNetV2~\cite{yu2020bisenet}, HarDNet~\cite{chao2019hardnet}, ICNet~\cite{zhao2018icnet}, DeepLabV3~\cite{chen2017rethinking}, GCNet~\cite{cao2019gcnet}, GSCNN~\cite{takikawa2019gated}, OCRNet~\cite{yuan2019object}, and so on. The number of segmentation models in PaddleSeg continues to increase in a period of time. Currently, there are 50+ pre-trained models on both Cityscapes~\cite{cordts2016cityscapes} and Pascal VOC 2012~\cite{everingham2010pascal}, which are most popular datasets in image segmentation. Later, we will provide comprehensive benchmarks and evaluations to show that the accuracy of these models in PaddleSeg outperforms other implementations. Besides the segmentation models in academic research, we also provide rich practical applications in PaddleSeg, which can help the developers to quickly build hands-on experiences on the real image segmentation cases.

The remainder of the paper is organized as follows. Section 2 presents the overview of PaddleSeg, and the major features. Section 3 describes the design strategies of high-quality models in PaddleSeg, and the evaluations of the models. Section 4 presents a conclusion.

\subsection{Contributions}

\begin{itemize}
  \item A high-efficient development toolkit for image segmentation that helps both developers and researchers in the whole process of designing segmentation models, training models, optimizing performance and inference speed, and deploying models.
  \item A lot of well-trained segmentation models and various real-world applications in both industry and academia, that help users conveniently build hands-on experiences in image segmentation.
  \item The high-quality segmentation models in PaddleSeg achieves higher accuracy than other implementations, which are beneficial to downstream algorithms and applications.
  
\end{itemize}

\section{PaddleSeg Overview}

\begin{figure}[t]
\begin{center}
\includegraphics[width=6.5in]{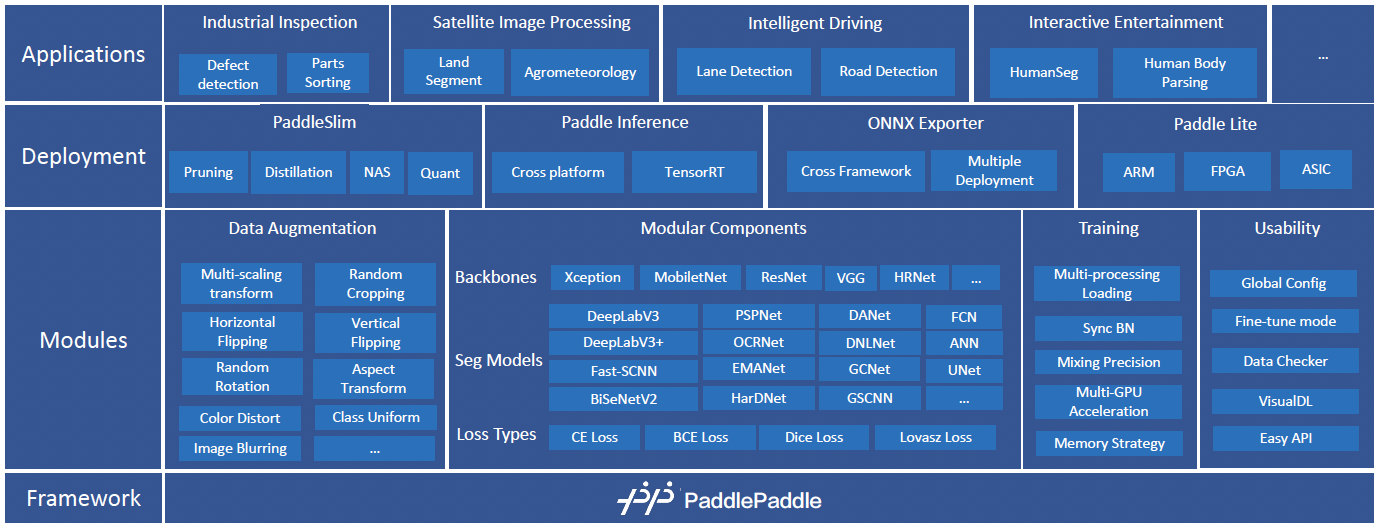}
\end{center}
\caption{The Overview of PaddleSeg Toolkit.}
\label{fig:seg_overview}
\end{figure}

In the section, we will give an overview of PaddleSeg as shown in Figure~\ref{fig:seg_overview}. PaddleSeg is built on PaddlePaddle\footnote{https://github.com/PaddlePaddle/Paddle}, which is a high-performance machine learning framework supporting ultra-large-scale training of deep neural networks~\cite{ma2019paddlepaddle}.

\subsection{Core Modules}

Upon the core framework, we build fundamental modules that are the most important for image segmentation, such as data augmentation, modular components, training optimization, and usability. Initially, PaddleSeg is designed to be an open-source project, which allows the developers to build their own segmentation models easily, and thus we provide a lot of image methods on data augmentation and configurable components. The training performance is a strength of PaddlePaddle, and PaddleSeg has developed further training optimization strategies, e.g. it supports multi-process data loading and training to speed up the whole training process. Also, PaddleSeg has enabled the feature of batch normalization synchronization across the multiple GPUs. As for alternative practical features, e.g. mixing-precision training, multi-GPU acceleration, and GPU memory optimization strategy, are also available in PaddleSeg.

In order to advance the user experience, PaddleSeg allows the developers to run the model in a global configuration mode, where they do not have to touch any code of segmentation models but only modify few values in a YAML file. This feature is quite convenient for entry-level developers who want to validate or train a model quickly. If the developers want to use our pre-trained models and fine-tune on a new dataset, we also provide the fine-tune mode for them. Also, dataset format or configuration is a common issue in image segmentation, and it is hard to find out the errors for most developers. PaddleSeg provides a data checker that helps them to locate the errors before the training process. While training the model, it is important to know how it works and the metrics 
fluctuate during the training. VisualDL~\cite{visualdl} in PaddleSeg provides a variety of charts to show the trends of parameters, visualizes model structures, and tracks the metrics in real-time. With VisualDL, the developers could understand the training process and the model structure more clearly. For those developers who only need to use well-built models or libraries, e.g. developing an image segmentation application, PaddleSeg provides easy APIs, in which the developers import paddleseg package using the pip install command. 
In short, the target users of PaddleSeg could cover all levels of developers.

On the fundamental modules, PaddleSeg has properly integrated other powerful toolkits to help developers further optimize the models and deploy the well-trained models into multiple types of devices. PaddleSlim is a toolkit for model compression. It contains a collection of compression strategies, such as pruning, fixed-point quantization, knowledge distillation, hyperparameter tuning, and neural architecture search. Paddle Inference is a high-performance inference engine for server-side applications, which achieves high throughput and low latency. ONNX Exporter is a toolkit which converts trained models of PaddleSeg to ONNX format so that the converted models can run on any device supported by the ONNX community. Paddle Lite is a light-weight inference engine for mobile, embedded, and IoT devices. It supports a diversity of hardware with high performance and high compatibilities, such as ARM CPU, Mali GPU, Adreno GPU, Huawei NPU, and FPGA. With the integration of powerful toolkits in PaddleSeg, the developers can effectively optimize the training model and deploy the model on multi-sided devices.

In addition, PaddleSeg provides various applications in image segmentation which are quite broad spanning industrial inspection, satellite image processing, intelligent driving, and interactive entertainment. In these applications, we provide detailed tutorials to show how to prepare the dataset, design a good model for specific segmentation problems, and optimize the performance of a trained model using post-processing strategies. These applications in practice could help the developers build more hands-on experiences in image segmentation. From the high-quality models to real-world applications, they are able to obtain the design insights and know-how easily. To the best of our knowledge, in the image segmentation field, PaddleSeg is the only one toolkit that supports such an end-to-end development process including data preparation, model design, model training, model optimization, model deployment, and real-world applications in both academia and industry.

\subsection{Major features}

The overview of PaddleSeg shows a bunch of good features, and we summarize the four most significant features from these.

\begin{figure}[t]
\begin{center}
\includegraphics[width=6.5in]{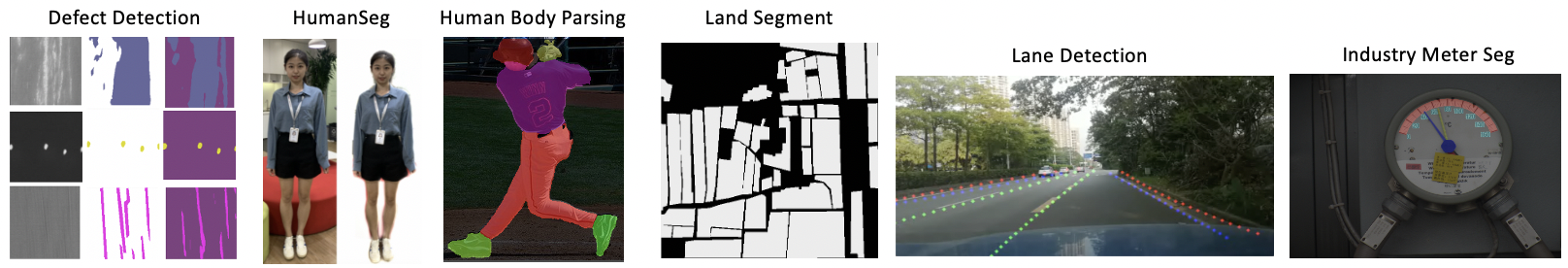}
\end{center}
\caption{The Examples of Applications.}
\label{fig:seg_cases}
\end{figure}

\begin{itemize}
  \item \textbf{Modular Design.} It is a core feature throughout the whole design of components. In PaddleSeg, we design five types of component managers to contain the models, backbones, losses, transformations, datasets, respectively. The benefit of such a design is that the users could construct the segmentation models according to their specific datasets and applications. For example, if you want to build an application on mobile devices, you had better choose a light-weight backbone, such as MobileNet. If you want to achieve a high-accuracy model running on GPUs, you could use a strong backbone, such as ResNet, or HRNet. For some particular practice, e.g. tiny object, or class imbalance, you can configure different losses and transformations components to tackle it efficiently.
  \item \textbf{High-Quality Models.} In PaddleSeg, we have applied a powerful distillation technique named semi-supervised label knowledge distillation solution (SSLD~\cite{ssld}), which helps the backbone networks have an improvement of more than 3\%. Base on the high-accuracy backbones, we provide around twenty high-quality models which are competitive on two popular datasets (Cityscapes, and Pascal VOC 2012) in image segmentation. 
  \item \textbf{Industry-level Deployment.} With the integration of four toolkits, i.e. PaddleSlim, Paddle Inference, ONNX Exporter, and Paddle Lite, PaddleSeg helps the developers optimize their model design and optimize model deployment, so that the well-trained models are able to run efficiently on multi-side devices, such as a server, mobile, and embedded devices.
  \item \textbf{Various Practical Cases.} PaddleSeg has completed various segmentation tasks with internal and external collaborators, and they have been becoming open-sourced to the developers already. We released six cases in the toolkit as shown in Figure~\ref{fig:seg_cases}, defect detection, human segmentation, human body parsing, land segment, lane detection, and industry meter segmentation. Besides, we provide detailed tutorials in the toolkit so that the developers are able to run these cases easily.
  
\end{itemize}

\section{High-Quality Models in PaddleSeg}

In the above section, we have already mentioned the high-quality models as one of the major features in PaddleSeg. In this section, we will discuss more details on the design and performance of these models.

\subsection{Strategies of Segmentation Models}

\begin{figure}[t]
\centering
{\subfloat[The segmentation network]{\includegraphics[width=0.5\columnwidth]{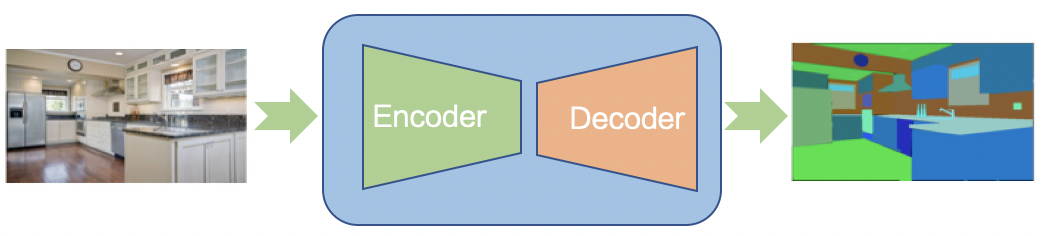}{\label{fig:seg_task}}}
\hfill
 \subfloat[The Encoder-Decoder Architecture~\cite{badrinarayanan2017segnet}]{\includegraphics[width=0.45\columnwidth, height=1in]{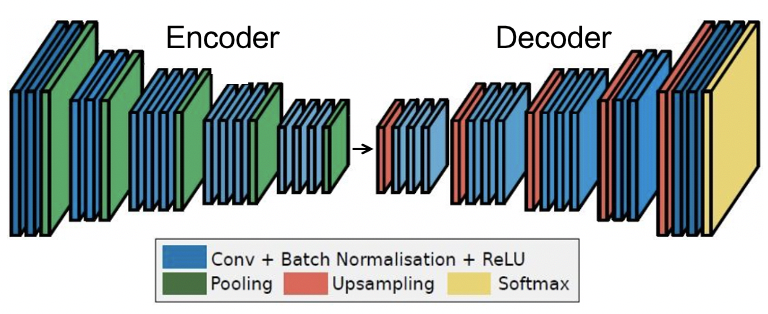}{\label{fig:encoder_decoder}}}
 }
 \caption{The segmentation network}
\label{fig:seg_network}
\end{figure}

First of all, we take a look at a generic model for the image segmentation task. As shown in Figure~\ref{fig:seg_task}, it consists of three parts, an input image, a segmentation network, and an output result. Different from classification and detection tasks, the size of the input image is the same as the size of the input in the segmentation task, and each pixel in the input image is assigned with a class label (different colors in the pseudo-color image). The segmentation network is composed of encoder and decoder. In the encoder part, the size of the feature maps is decreasing gradually. On the contrary, the size of feature maps is increasing to the same size of the input image in the decoder. Figure~\ref{fig:encoder_decoder} indicates a encoder-decoder architecture with more details. In theory, this pure encoder-decoder model is able to almost complete an image segmentation task properly, while the segmentation results are usually far from expectations~\cite{long2015fully}. Therefore, besides the generic encoder-decoder model, there are more useful approaches that help to improve the accuracy in a high-quality segmentation model, such as multiple scales~\cite{tao2020hierarchical,chen2016attention}, boundary refinement~\cite{yuan2020segfix,takikawa2019gated}, auxiliary loss~\cite{zhao2017pyramid, yuan2019object, tao2020hierarchical}, and so on. From our experience of the development in PaddleSeg, we summarize five crucial types of strategies and then discuss them in the following.

\textbf{Skip Connection.} The drawback of a simple encoder-decoder model is obvious: as the resolution of the input image is decreasing, the low-level detailed information of the input is lost, and eventually the encoder outputs high-level information of the input. The decoder only receives the high-level features, resulting in the coarse segmentation results. Therefore, it is necessary to combine low-level and high-level features in the decoder by the skip connection, and then the decoder can obtain the detailed information of the input by enhancing the features of different levels. The Fully Convolutional Network (FCN)~\cite{long2015fully} pioneers the skip-connection strategy, and U-Net~\cite{ronneberger2015u} push the skip connection further, where it utilizes the symmetrical encoder-decoder architecture. Different from FCN, U-Net fully leverages the features from each layer by using dense skip connections. The features from each layer in the encoder part are connected to the symmetrical layers in the decoder part. U-Net has drawn extensive attention from the medical image analysis community~\cite{hao2020brief}. Besides U-Net, there are several similar networks implementations in PaddleSeg, such as Attention U-Net~\cite{oktay2018attention}, U-Net++~\cite{zhou2018unet++}, U\textsuperscript{2}-Net~\cite{qin2020u2}.

\textbf{Dilated Convolution.} As discussed above, the resolution is decreasing gradually by downsampling and results in information lost. In order to keep the resolution, we can reduce the number of down-sampling layers. However, such a method would bring another problem that its computational complexity is increasing dramatically and the size receptive field is narrowing. The segmentation result is not very accurate either. The better approach is introducing the dilated convolution, which enlarges the receptive field by keeping the resolution and computational complexity. In a dilated convolutional layer, a 3x3 kernel with a dilation rate of 2 will have
the same size receptive field as a 5x5 kernel while using
only 9 parameters, but a normal 5x5 kernel without the dilated rate needs 25 parameters. Therefore, it will capture more complete features with the same size of convolution. Dilated convolutions have been widely used in the field of real-time segmentation, and many recent publications report the use of this technique, where the DeepLab series are the most representative work~\cite{minaee2020image}. In PaddleSeg, we have already implemented DeepLabV3~\cite{chen2017rethinking} and DeepLabV3+~\cite{chen2018encoder}, which are popular in the many applications.

\textbf{Global Context.} Besides dilated convolutions, there is an alternative way to increase the receptive field, i.e. leveraging context information. Since the convolution is a local operation in nature, it tends to cause a lack of context. In general, the context information, which is far beyond the pixel-level appearance, becomes aware of semantics, and provides a useful complementary source for building segmentation models~\cite{hao2020brief}. Pyramid pooling was initially proposed in the image classification and object detection~\cite{he2015spatial}. Then it has been introduced in image segmentation to obtain context information of different scales, which can enlarge the receptive field, and finally combine local and global information to assist decision-making. For example, ships in the water will be probably classified as cars in the local view, since the local appearance of the ship is quite similar to the car. Combined with the global context information, they can be correctly classified, because the model has already known the context that the target is on the water. The most famous context-based methods are pyramid pooling (PP) module of PSPNet~\cite{zhao2017pyramid}, as well as atrous spatial pyramid pooling (ASPP) module of DeepLab family~\cite{chen2017deeplab,chen2017rethinking,chen2018encoder}, where The PP module and ASPP module have already been implemented in PaddleSeg. Apart from local features, the global context also helps to effectively produce more accurate and smooth segmentation results.

\textbf{Attention Mechanism.} In addition to context information, the attention mechanism emphasizes the relationship modeling between pixels. Since the convolution is a block operation, it is hard to establish the relationship between all pixels. However, the importance of all pixels is not equal for a pixel to be classified. The attention
mechanism captures long-range dependencies in an effective manner by allowing the model to automatically search for pixels that are relevant to classifying a target pixel. If the importance of the relationship between pixels can be modeling, it can obviously enhance the contribution of pixels from the same class of objects, and obtain a better segmentation result. while classifying a pixel inside an object, it will be more inclined to assist the classification through internal pixels belonging to the same object with the target pixel~\cite{yuan2019object}. The attention mechanism was initially proposed in the field of machine translation, and has been persistently explored in image segmentation already. PaddleSeg has implemented several methods using the attention mechanism. The representatives here are DANet~\cite{fu2019dual}, which adopts dual attention mechanism, and OCRNet~\cite{yuan2019object}, which captures object-contextual representations. Multi-scale attention mechanism softly weights the multi-scale features at each pixel location~\cite{tao2020hierarchical}.

\textbf{Strong Backbone.} Since the image segmentation task can be regarded as a pixel classification problem, the accuracy of backbone networks influences the segmentation performance indirectly. In general, high-accuracy segmentation models usually utilize strong backbone networks. In PaddleSeg, we have implemented several backbone networks commonly used in image segmentation. VGG~\cite{simonyan2014very} uses multiple 3 ×3 convolutions in the sequence that can match the effect of larger receptive fields, e.g. 5 ×5 and 7 ×7. MobileNet and its variants~\cite{howard2017mobilenets,sandler2018mobilenetv2,howard2019searching} introduce depthwise convolution layers that achieve a great balance between accuracy and computational cost. Xception~\cite{chollet2017xception} replaces the inception modules with depthwise separable convolutions, which has better performance in image classification, and also demonstrates its efficiency in image segmentation in DeepLabV3+. ResNet~\cite{he2016deep} successfully enables a much deeper network and models the residual representation into the CNN network structure, which solves the difficulty of training a very deep network structure. HRNet~\cite{wang2020deep} maintains high-resolution representations through the encoding process by connecting the high-to-low resolution convolution streams in parallel. Many of the more recent works on segmentation use ResNet and HRNet as the backbone, due to their high accuracy performance. In PaddleSeg, with the benefits of our knowledge distillation solution, these backbone networks achieve much higher accuracy and also improve segmentation results.

\subsection{Analysis of High-Quality Models}

As mention above, a high-quality segmentation model requires several strategies to improve its accuracy, and also smooth the boundary. Currently, PaddleSeg has implemented a lot of representative models that possess one or more strategies as shown in Table~\ref{table:models}. Skip connection and backbone are the most two strategies among these models, which have been fully proved to be effective in image segmentation. If a model only has one or two strategies, it could be a light-weight model for real-time segmentation applications, such as BiSeNetV2, Fast SCNN, HarDNet, and ICNet. U-Net is designed for medical imaging applications so that it does not share common properties with other models. Since FCN is using HRNet as the backbone network, it applies a simple segmentation head upon the output of the backbone. Among these models, there are three models that have the most strategies, i.e. DeepLabV3+, GSCNN, and OCRNet. Due to the high accuracy and efficiency, DeepLabV3+ is widely used in both industrial and mobile applications. GSCNN is a recent model with edge and shape constraints resulting in a better boundary segmentation. OCRNet and its extensions have yielded SOTA results on various datasets of image segmentation.

\begin{table}[h!]
\centering
\begin{tabular}{ |c|c|c|c|c|c| } 
 \hline
 \textbf{Models} & \textbf{Skip Connection} & \textbf{Dilated Conv} & \textbf{Context} & \textbf{Attention} & \textbf{Backbone} \\ 
 \hline
 ANN~\cite{zhu2019asymmetric} & \checkmark &  &  & \checkmark & \checkmark \\ 
 \hline
 BiSeNetV2~\cite{yu2020bisenet} & \checkmark &   & \checkmark &  &   \\ 
 \hline
  DANet~\cite{fu2019dual} & \checkmark &   &   & \checkmark & \checkmark  \\ 
 \hline
  DeepLabV3~\cite{chen2017rethinking} &   & \checkmark & \checkmark &  & \checkmark  \\ 
 \hline
  DeepLabV3+~\cite{chen2018encoder} & \checkmark & \checkmark & \checkmark &  & \checkmark  \\ 
 \hline
   DNL~\cite{yin2020disentangled} & \checkmark &  &  & \checkmark & \checkmark  \\ 
 \hline
  EMANet~\cite{li2019expectation} & \checkmark &  &  & \checkmark & \checkmark  \\ 
 \hline
  FastSCNN~\cite{poudel2019fast} & \checkmark &  & \checkmark &  & \checkmark  \\ 
 \hline
   FCN~\cite{wang2020deep} &   &  &  &  & \checkmark  \\ 
 \hline
  GCNet~\cite{cao2019gcnet} & \checkmark &  &  & \checkmark & \checkmark  \\ 
 \hline
  GSCNN~\cite{takikawa2019gated} & \checkmark & \checkmark & \checkmark &  & \checkmark  \\ 
 \hline
   HarDNet~\cite{chao2019hardnet} & \checkmark &  &  &  &   \\ 
 \hline
  ICNet~\cite{zhao2018icnet} & \checkmark &  &  &  &   \\ 
 \hline
  OCRNet~\cite{yuan2019object} & \checkmark &  & \checkmark & \checkmark & \checkmark  \\ 
 \hline
  PSPNet~\cite{zhao2017pyramid} & \checkmark &  & \checkmark &  & \checkmark  \\ 
 \hline
  U-Net~\cite{ronneberger2015u} & \checkmark &  &  &  &   \\ 
 \hline
 \end{tabular}
 \vspace{0.1in}
\caption{Analysis of high-quality segmentation models implemented in PaddleSeg. Models are sorted by the initial. We summarize 5 crucial types of strategies for high-quality models, i.e. skip connection, dilated convolution, context, attention mechanism and strong backbone. Models are sorted by the initial. If the model has a backbone, the first 4 strategies are taken into consideration only in segmentation head.}
\label{table:models}
\end{table}

We have conducted the empirical analysis experiments of these models on Cityscapes validation set. The Cityscapes dataset~\cite{cordts2016cityscapes} is tasked for urban scene understanding, where it provides dense pixel-level annotations for 5000 images at 2048 x 1024 resolution pre-split into training (2975), validation (500), and test (1525) sets. There are totally 30 classes and only 19 classes are used for parsing evaluation.

We set the initial learning rate as 0.01, weight decay as 4e-5, crop
size as 1024 × 512 and batch size as 8 by default for most models, but there are different configurations for some particular models, and all detailed configurations and documents can be found in PaddleSeg. In addition to training configurations, we also provide all training logs, well-trained weights, and VisualDL logs so that the developers are able to observe our training process clearly, and reproduce the results easily.

In Table~\ref{table:resnet101}, we compare the performance of above models. DeepLabV3+ and OCRNet with even different backbone networks consistently perform better than other methods. We only conducted the experiment on GSCNN with ResNet50, because it cannot maintain a fair configuration with ResNet101. This experiment result is consistent with our above analysis that the strategies are crucial to achieving the high-accuracy models, and the three models have the most ones. It is worth noting that HRNet even with a simple FCN head outperforms some well-designed models, which demonstrates the importance of a strong backbone network. In our implementations, the performance improves further with knowledge distillation on the backbones. For the light-weight models, BiSeNetV2 and HarDNet are potentially useful for real-time applications, where they are reported to achieve over 50 fps in the original articles. Due to its simplicity and friendliness to mobile devices,  Fast SCNN is quite popular in the industry, though the accuracy is lower than BiSeNetv2 and HarDNet. 

We have analyzed a lot of characteristics of high-accuracy models, as well as their methods and strategies. In fact, the modular design of PaddleSeg allows the developers to conveniently apply efficient strategies, which are off-the-shelf components. There are possible solutions that the developers can combine different components to achieve a higher-accuracy model. We could also conduct more ablation studies on such combinations in future work, to demonstrate the effects of different components.

\begin{table}[t]
\begin{tabular}{ll}
\centering
 \hspace{0.5in}
\begin{tabular}{ |c|c|c| } 
 \hline
 \textbf{Models} & \textbf{Backbone} & \textbf{mIoU} \\ 
  
  \hline
  EMANet & ResNet50 & 77.58  \\ 
   \hline
  PSPNet & ResNet50 & 78.83 \\ 
   \hline
   FCN & HRNet\_w18  & 78.97  \\ 
 \hline
 ANN & ResNet50 & 79.09 \\ 
 \hline
   DNL & ResNet50 & 79.28  \\ 
   \hline
  GCNet & ResNet50 & 79.50 \\ 
   \hline
    DeepLabV3 & ResNet50  & 79.90 \\ 
 \hline
  DANet & ResNet50 & 80.27\\

 \hline
  DeepLabV3+ & ResNet50 &  80.36\\ 

 \hline
  GSCNN & ResNet50 & 80.67  \\ 

 \hline
  OCRNet & HRNet\_w18 &  80.67 \\ 
  \hline
  
 \end{tabular}\hfill
 \hspace{0.5in}
 \begin{tabular}{ |c|c|c| } 
 \hline
 \textbf{Models} & \textbf{Backbone} & \textbf{mIoU} \\ 
 \hline
  U-Net & - &  66.89 \\ 
   \hline
  ICNet & - &  68.31 \\ 
   \hline
  FastSCNN & - & 69.31 \\ 
   \hline
 BiSeNetV2 & - & 73.19\\ 
    \hline
   HarDNet & - &  79.03 \\ 
   \hline
 EMANet & ResNet101 & 79.42\\ 
  \hline
  PSPNet & ResNet101 & 80.48  \\ 
   \hline
  ANN & ResNet101 & 80.61 \\ 
   \hline
   FCN & HRNet\_w48  & 80.70  \\ 
   \hline
   DeepLabV3 & ResNet101 &  80.85\\ 
 \hline
 GCNet & ResNet101 & 81.01 \\ 
 \hline
   DeepLabV3+ & ResNet101 & 81.10  \\ 
   \hline
  OCRNet & HRNet\_w48 & 82.15 \\ 
\hline
 \end{tabular}

\end{tabular}
 \vspace{0.1in}
 \caption{Results on Cityscapes val. The crop size is set as 1024 x 512 by default, except the models without backbone, where their crop size is set as 1024 x 1024. The segmentation accuracy is reported in terms of mean IoU (\%).}
\label{table:resnet101}
\end{table}

\subsection{Comparison with Other Implementations}

To compare model performance in PaddleSeg with other implementations, we carry out detailed experiments on two mainstream segmentation datasets: Cityscapes~\cite{cordts2016cityscapes} and PASCAL VOC 2012~\cite{everingham2010pascal}. PASCAL VOC 2012 consists of 1464, 1449, and 1456 images for training, validation, and testing, respectively. It includes 20 object classes and one background class. In our experiment, we use an augmented version~\cite{hariharan2011semantic} of PASCAL VOC 2012 which has 10,582 images in the training set. 

\textbf{Training Settings.} The backbone networks are pre-trained on the ImageNet~\cite{deng2009imagenet}. All models are using a unified configuration as follows. We use Stochastic Gradient Descent (SGD) to optimize our network, where we set the initial learning rate to 0.01, and set the momentum to 0.9 with weight decay 4e-5 for both datasets. During training, the learning rate is decayed according to the “poly” leaning rate policy, where the learning rate is multiplied by $1-\left ( \frac{iter}{max\_iter} \right )^{power}$ where power = 0.9.
For both datasets, we employ random scaling in the range of [0.5, 2.0],
random horizontal flip, and random brightness as additional
data augmentation methods. We use a crop size of 1024x512 for Cityscapes and 512x512 for PASCAL VOC. The batch size is 8 in Cityscapes
experiments and 16 in the PASCAL VOC. Synchronized the batch normalization is enabled to synchronize the mean and standard-deviation of batch normalization layer across multiple GPUs. We train on the training set of Cityscapes, and PASCAL VOC for 80K, 40K iterations, respectively.  There is one exception though: OCRNet trained on Cityscapes for 160K iterations for a fair comparison. All training settings including detailed training logs are available in the PaddleSeg repository.

In Table~\ref{table:cityscapes}, we show the performance evaluation on the Cityscapes validation set. Compared to the results of other implementations, the accuracy of models in PaddleSeg has the obvious improvement, especially ANN, GCNet and OCRNet which are based on the attention mechanism. Moreover, the accuracy improvement on PASCAL VOC is even more dominant shown in Table~\ref{table:voc}, up to 4\% gain, where results of all models are beating other implementations. The comprehensive comparison demonstrates that the high-accuracy segmentation algorithm in PaddleSeg is very competitive.

\begin{table}[t]
\begin{tabular}{ll}
\centering
 \hspace{0.3in}
\begin{tabular}{ |c|c|c|c| } 
 \hline
 \textbf{Models} & \textbf{Others} & \textbf{PaddleSeg}& \textbf{Gain} \\ 
  \hline
  ANN & 77.34 & 79.09 & 1.75 $\uparrow$ \\ 
   \hline
  DANet & 79.34 & 80.27 & 0.93 $\uparrow$\\ 
   \hline
   DeepLabV3 & 79.32  & 79.90 & 0.58 $\uparrow$ \\ 
 \hline
 DeepLabV3P & 80.09 & 80.36 & 0.27 $\uparrow$\\ 
  \hline
 EMANet & 77.59 & 77.58 & - \\ 
 \hline
   FCN & 78.80 & 78.97  & 0.17 $\uparrow$\\ 
   \hline
  GCNet & 78.48 & 79.50 & 1.02 $\uparrow$\\ 
   \hline
    OCRNet & 79.47  & 80.67 & 1.20 $\uparrow$\\ 
 \hline
  PSPNet & 78.55 & 78.83 & 0.28 $\uparrow$\\
 \hline
  
 \end{tabular}\hfill
 \hspace{0.3in}
 \begin{tabular}{ |c|c|c|c| } 
 \hline
 \textbf{Models} & \textbf{Others} & \textbf{PaddleSeg}& \textbf{Gain} \\ 
 
 \hline
  ANN & 78.80 & 80.61 & 1.81 $\uparrow$  \\ 
   \hline
   DeepLabV3 & 80.20  & 80.85 & 0.65 $\uparrow$  \\ 
 \hline
 DeepLabV3P & 80.97 & 81.10 & 0.13 $\uparrow$ \\ 
   \hline
 EMANet & 79.10 & 79.42 & 0.32 $\uparrow$ \\ 
 \hline
   FCN & 80.65 & 80.70 & -  \\ 
   \hline
  GCNet & 79.03 & 81.01 & 1.98 $\uparrow$ \\ 
   \hline
    OCRNet & 81.35  & 82.15 & 0.80 $\uparrow$ \\ 
 \hline
  PSPNet & 79.76 & 80.48 & 0.72 $\uparrow$\\
 \hline
 \end{tabular}

\end{tabular}
 \vspace{0.1in}
 \caption{Comparison of results on Cityscapes val. The models in the left table use ResNet50 as the backbone, while FCN and OCRNet use HRNet\_w18. The models in the right table use ResNet101 as the backbone, while FCN and OCRNet use HRNet\_w48. The segmentation accuracy is reported in terms of mean IoU (\%). Ignore the gap if the difference is less than 0.1\%.}
\label{table:cityscapes}
\end{table}

\begin{table}[t]
\begin{tabular}{ll}
\centering
 \hspace{0.3in}
\begin{tabular}{ |c|c|c|c| } 
 \hline
 \textbf{Models} & \textbf{Others} & \textbf{PaddleSeg} & \textbf{Gain} \\ 
  \hline
  ANN & 76.56 & 80.82 & 4.26 $\uparrow$  \\ 
   \hline
  DANet & 76.37 & 78.55 & 2.18 $\uparrow$ \\ 
   \hline
   DeepLabV3 & 77.68  & 79.76 & 2.08 $\uparrow$  \\ 
 \hline
 DeepLabV3P & 76.81 & 80.66 & 3.85 $\uparrow$ \\ 
 \hline
   FCN & 72.90 & 75.39 & 2.49 $\uparrow$  \\ 
   \hline
  GCNet & 76.24 & 80.32 & 4.08 $\uparrow$ \\ 
   \hline
    OCRNet & 74.98  & 75.76 & 0.78 $\uparrow$ \\ 
 \hline
  PSPNet & 77.29 & 80.76 & 3.47 $\uparrow$\\
 \hline
  
 \end{tabular}\hfill
 \hspace{0.3in}
 \begin{tabular}{ |c|c|c|c| } 
 \hline
 \textbf{Models} & \textbf{Others} & \textbf{PaddleSeg} & \textbf{Gain} \\ 
 
 \hline
  ANN & 76.70 & 79.62 & 2.92 $\uparrow$  \\ 
   \hline
   DeepLabV3 & 77.92 & 80.62& 2.70 $\uparrow$  \\ 
 \hline
 DeepLabV3P & 78.62 & 80.60& 1.98 $\uparrow$ \\ 
 \hline
   FCN & 76.24 & 78.72 & 2.48 $\uparrow$  \\ 
   \hline
  GCNet & 77.84 & 79.64 & 1.80 $\uparrow$ \\ 
   \hline
    OCRNet & 77.14  & 79.98 & 2.84 $\uparrow$\\ 
 \hline
  PSPNet & 78.52 & 80.76 & 2.24 $\uparrow$\\
 \hline
 \end{tabular}

\end{tabular}
 \vspace{0.1in}
 \caption{Comparison of results on Pascal VOC val. The models in the left table use ResNet50 as the backbone, while FCN and OCRNet use HRNet\_w18. The models in the right table use ResNet101 as the backbone, while FCN and OCRNet use HRNet\_w48.}
\label{table:voc}
\end{table}

\section{Conclusion}

In this paper, we presented a high-efficient development toolkit for image segmentation called PaddleSeg, which helps both developers and researchers in the whole process of the development of state-of-the-art segmentation models. PaddleSeg supports up-to-date segmentation models with best practice covering both real-time and high-accuracy levels, and over fifty pre-trained models are provided in the repository. With the modular design, the users can conveniently build high-accuracy models by off-the-shelf components in PaddleSeg. The empirical analysis and comprehensive evaluations also show that these segmentation algorithms trained on our toolkit have more competitive accuracy. In the future, besides introducing more high-quality models and datasets, we will further explore the practical solutions in vertical fields with collaborators, such as medical imaging, remote sensing imaging, intelligent driving, and mobile applications.

\bibliographystyle{unsrt}  
\bibliography{references}  

\end{document}